# Disparities in Social Determinants among Performances of Mortality Prediction with Machine Learning for Sepsis Patients

Hanyin Wang [1], Yikuan Li [1], Andrew Naidech [2], Yuan Luo [1*]

Emails:

HanyinWang2022@u.northwestern.edu

Yikuan.Li@northwestern.edu

Andrew.Naidech@nm.org

Yuan.Luo@northwestern.edu

[1] Department of Preventive Medicine, Feinberg School of Medicine, Northwestern University, Chicago, Illinois, USA

[2] Department of Neurology, Feinberg School of Medicine, Northwestern University, Chicago, Illinois, USA

* Corresponding author

Abstract

*Background*

Sepsis is one of the most life-threatening circumstances for critically ill patients in the United States, while diagnosis of sepsis is challenging as a standardized criteria for sepsis identification is still under development. Disparities in social determinants of sepsis patients can interfere with the risk prediction performances using machine learning.

*Methods*

We analyzed a cohort of critical care patients from the Medical Information Mart for Intensive Care (MIMIC)-III database. Disparities in social determinants, including race, gender, marital status, insurance types and languages, among patients identified by six available sepsis criteria were revealed by forest plots with 95% confidence intervals. Sepsis patients were then identified by the Sepsis-3 criteria. Sixteen machine learning classifiers were trained to predict in-hospital mortality for sepsis patients on a training set constructed by random selection. The performance was measured by area under the receiver operating characteristic curve (AUC). The performance of the trained model was tested on the entire randomly conducted test set and each sub-population built based on each of the following social determinants: race, gender, marital status, insurance type, and language. The fluctuations in performances were further examined by permutation tests.

*Results*

We analyzed a total of 11,791 critical care patients from the MIMIC-III database. Within the population identified by each sepsis identification method, significant differences were observed among sub-populations regarding race, marital status, insurance type, and language. On the 5,783 sepsis patients identified by the Sepsis-3 criteria statistically significant performance decreases for mortality prediction were observed when applying the trained machine learning model on Asian and Hispanic patients, as well as the Spanish-speaking patients. With pairwise comparison, we detected performance discrepancies in mortality prediction between Asian and White patients, Asians and patients of other races, as well as English-speaking and Spanish-speaking patients.

*Conclusions*

Disparities in proportions of patients identified by various sepsis criteria were detected among the different social determinant groups. The performances of mortality prediction for sepsis patients can be compromised when applying a universally trained model for each subpopulation. To achieve accurate diagnosis, a versatile diagnostic system for sepsis is needed to overcome the social determinant disparities of patients.

Keywords: Sepsis, machine learning, social determinants, disparity, mortality prediction.

# Background

Sepsis, one of the most life-threatening circumstances for critically ill patients in the United States, is the culmination of complex interactions between the infecting microorganism and the host immune, inflammatory, and coagulation responses. [1, 2] Each year, more than 1.7 million adults in the United States develop sepsis, and approximately 270,000 die because of sepsis. The prevalence of sepsis is around one-third among hospitalized patients. [3] With a few identification methods currently available, a standardized criteria is still under development. [4]

Disparities in critical care can be induced by multi-factored causes. [5-8] Biases are observed in healthcare for patients from different social status groups. [9, 10] With more data-driven and artificial intelligence (AI) involved in healthcare, disparities among sub-populations are more frequently observed and attracted more attention. [11-14] Machine learning applications for computational phenotyping and risk prediction in healthcare are becoming more powerful with the development of electronic health records (EHRs). [15-22] Risk predictions for sepsis patients using machine learning techniques have been studied. [23-26] However, the discussions over how the disparities and biases interact with risk prediction models for sepsis patients remain undefined. In this study, we revealed the disparities in the proportions of sepsis in subpopulations of social determinants groups from a cohort of patients admitted for critical care services and examined the fluctuations in the performances of mortality prediction for subpopulations of sepsis patients when using machine learning classifiers.

# Methods

*Data*

Medical Information Mart in Intensive Care (MIMIC)-III v1.4 is an open-sourced large scale database of critical care patients with enriched features. [27] From a total of 23,620 intensive care unit (ICU) admission records, 11,791 patients with their initial admission records were identified and utilized in this study. Selection criteria were applied to filter out nonadults, patients with suspected infection more than 24 hours before the ICU admission or more than 24 hours after the ICU admission, patients with missing data, and patients admitted for cardiothoracic surgery services. The data selection algorithms were elaborated in a previous study. [4]

*Social determinants*

Five social determinants were studied, including race, gender, insurance type, marital status, and language. Race of all subjects was re-leveled into five categories, Asian, Black or African American, Hispanic or Latino, White and other, where the "other" category covers American Indian and Alaska Native, Native Hawaiian or other Pacific Islander, multi-race, unspecified race, and other races not mentioned above. Dichotomous gender, female and male, was considered. Insurance types were taken directly from the MIMIC-III database, which includes government, Medicaid, Medicare, private, and self-pay. Marital status was re-factorized into the following categories: significant other, single, separated, widowed, and unknown, where the "significant other" category covers the situations if life partner or married was indicated in the MIMIC-III database, the "separated" category covers the circumstances if divorced or separated was displayed in the database, the "unknown" category covers the situation if unknown (default) was indicated in the database and was coded for those patients did not specify the marital status. Languages were re-grouped into English, Spanish and other, where the "other" category covers any languages documented in the database other than the two stated.

*Disparities in social determinants across various sepsis criteria*

We compared the disparities between each sub-category of social determinants in the sepsis population detected by the six identification methods for sepsis: (1) explicit criteria: two codes explicitly mentioning sepsis (995.92 and 785.52 for severe sepsis and septic shock, respectively) defined by International Classification of Diseases, 9$^{th}$ version (ICD-9); (2) Angus methodology [28]; (3) Martin methodology [29]; (4) criteria presented by Centers for Medicare & Medicaid Services (CMS) [30]; (5) the complete surveillance criteria presented by Center of Disease Control and Prevention (CDC) [31]; (6) Sepsis-3 [32]. Forest plots were generated for the proportion of each subpopulation that was identified as sepsis by each method. For example, a proportion of 0.274 for Asian and "Angus" represents 27.4% of the Asians in the dataset were identified as sepsis by the Angus criteria. A 95% confidence interval was constructed by bootstrapping (1,000 simulations) and shown in the forest plots for each proportion.

*Mortality prediction for sepsis patients using machine learning*

We built machine learning classifiers to predict mortality for sepsis patients. The sepsis patient population was constructed using the Sepsis-3 identification method since it is the latest and most conservative among the six methods being discussed. [4] The entire cohort of patients was split into training and testing sets to a proportion of 7:3. Sixteen machine learning configurations were built and trained to predict in-hospital mortality for the sepsis patients, that include Ridge classifier, perceptron, passive-aggressive classifier, k-nearest neighbors (kNN), random forest, support vector machine with linear kernel (linearSVC) and L1 or L2 regularization, support vector machine with linear kernel and L2 regularization, stochastic gradient descent (SGD) classifier with L1, L2, or elastic net regularization, multinomial naïve Bayes, Bernoulli naïve Bayes, logistic regression, support vector machine

(SVM) with rbf, polynomial, or sigmoid kernel. Sequential organ failure assessment (SOFA) score [33] during the first 24 hours of admission, systemic inflammatory response syndrome (SIRS) score [34] during the first 24 hours of admission, and age were employed as features. Before training the machine learning configurations, each feature was scaled to 0 to 1 to avoid the impact of different magnitudes. Five-fold cross-validation was employed to find the optimal hyper-parameters for each machine learning configuration. The best-suited thresholds for each classifier were set according to Youden's J statistics. The performances of the machine learning configurations were measured by the area under the receiver operating characteristic curve (AUC).

*Statistical analysis for disparities in performances on sub-populations of social determinants*
The training procedures were carried out on the entire training set, after which trained configurations and evaluation metrics on the entire cohort were saved. In the next step, we tested the performance on every sub-population of each of the five social determinants. To detect the disparities in performances, we compared the AUCs on the entire cohort with those on the subpopulations by permutation tests (1,000 times). A one-tailed permutation test was employed to determine if the decrease or increase of the performance is significant statistically when testing on sub-groups of patients. To further illustrate the disparities, we conducted pairwise permutation tests (1,000 times) among each pair of the sub-populations. A two-tailed permutation test was used to show if there are significant disparities in performances among each pair. The entire workflow can be found in Figure 1.

The analysis was conducted using Python 3.6.8. Machine learning classifiers, cross-validation, and evaluation metrics were conducted using Sci-kit Learn 0.23.2.

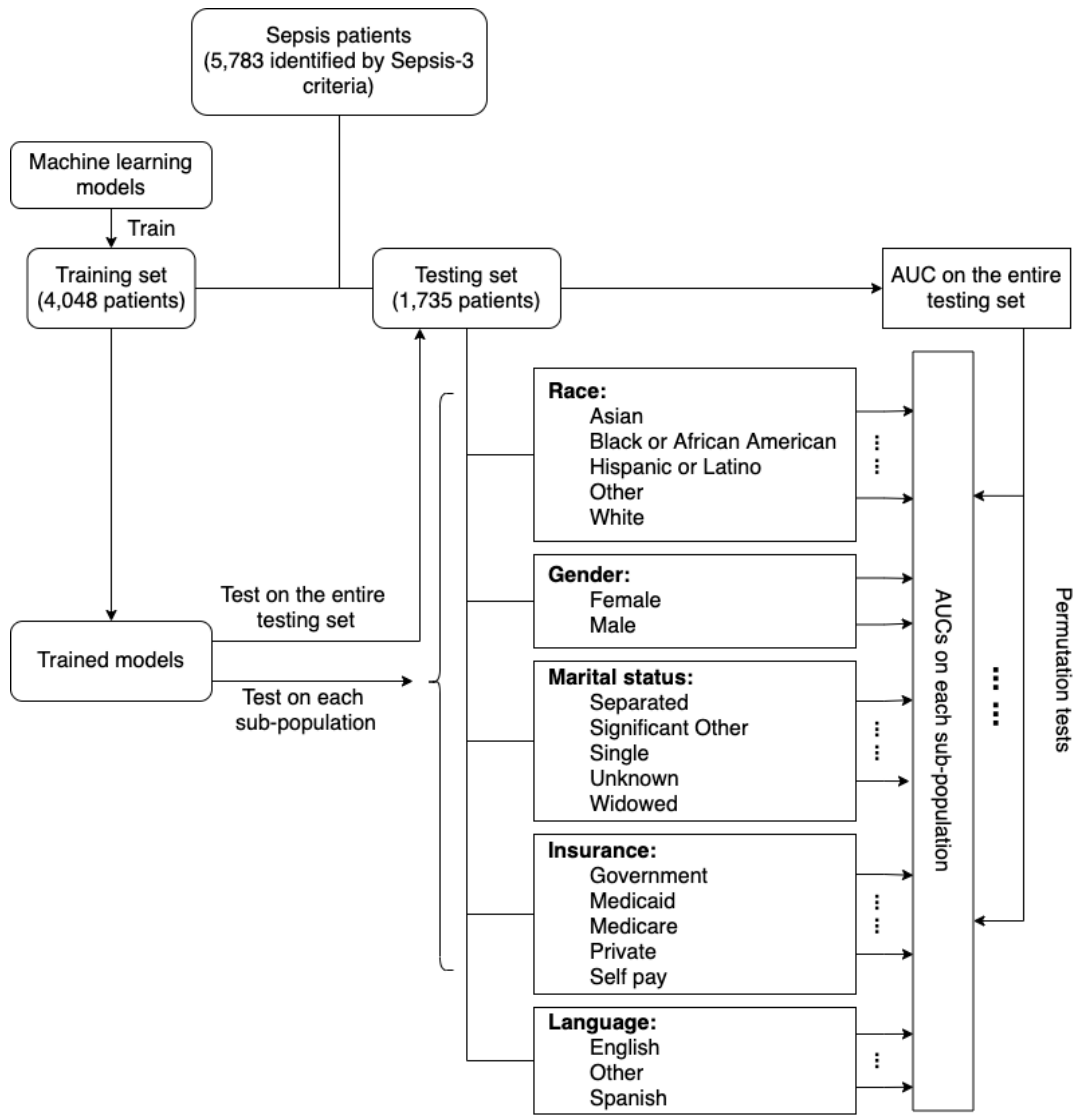

Figure 1. The workflow.

Results

*Disparities in social determinants across various sepsis criteria*

Forest plots for the disparities in social determinants across various sepsis criteria are shown in Figure 2. Proportions of sepsis patients identified by different methods showed significant discrepancies, with the Sepsis-3 criteria as the most conservative one. Within the population identified by the same sepsis identification method, significant differences were observed among sub-populations regarding race, marital status, insurance type, and language. Numeric values of the proportions and 95% confidence interval can be found in the Supplementary Table 1 in the supplementary information.

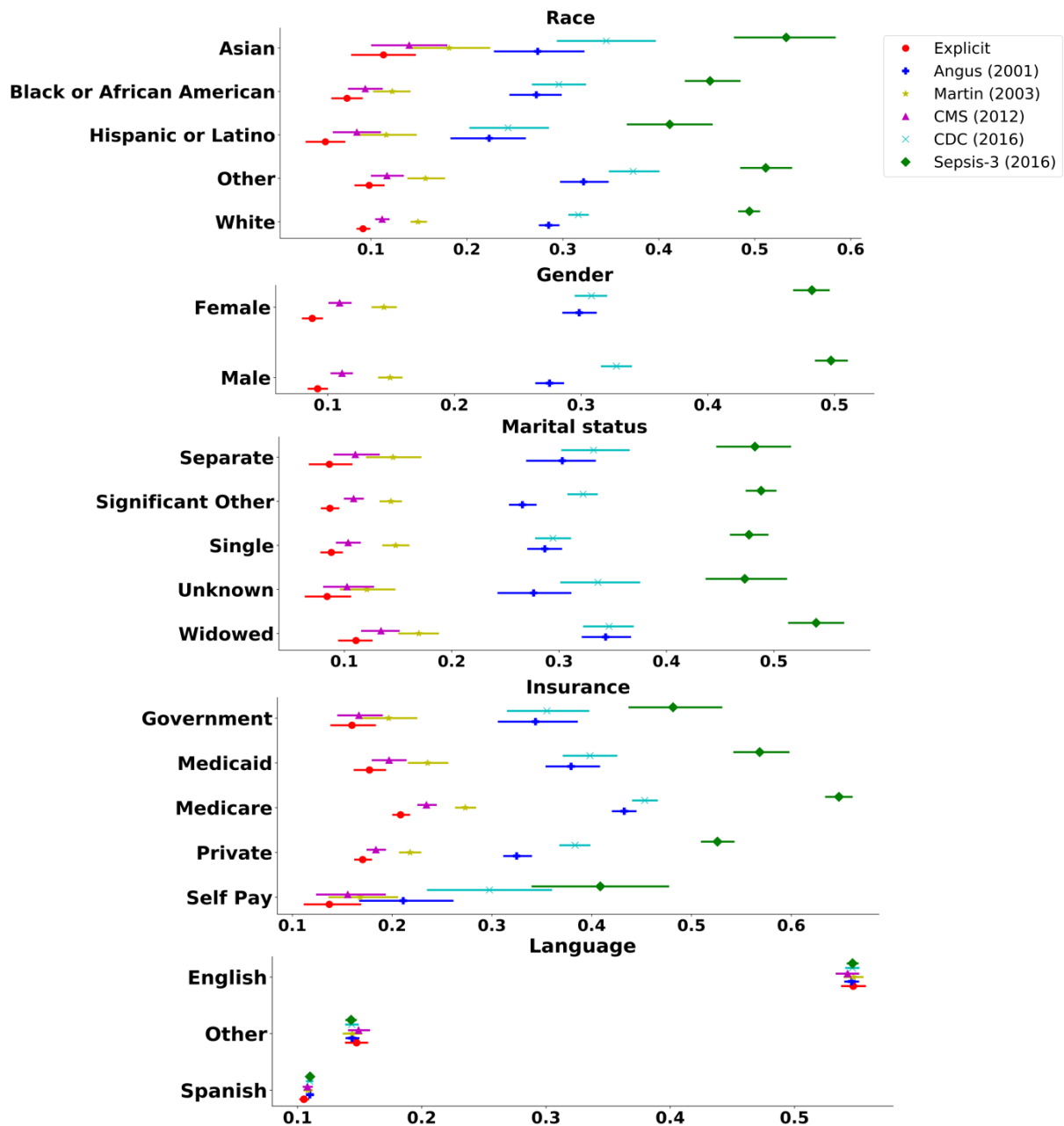

Figure 2. Forest plot for disparities in social determinants across various sepsis criteria. The proportions of sepsis patients of every sub-population identified by each sepsis criteria are shown as a point with a 95% confidence interval. Sepsis criteria are shown in different colors, while results for each subpopulation are shown in a row corresponding to the labels on the y-axis. Explicit: the explicit criteria; Angus: the Angus methodology; Martin: the Martine methodology; CMS: criteria presented by Centers for Medicare & Medicaid Services (CMS); CDC: the complete surveillance criteria presented by Center of Disease Control and Prevention (CDC); Sepsis-3: the Sepsis-3 criteria.

*Mortality prediction for sepsis patients using machine learning*

In total, 5,783 patients were identified as sepsis by the Sepsis-3 criteria. Statistics of this cohort of sepsis patients are shown in Table 1. The detailed testing performances on the entire testing set are shown in Table 2.

We compared the performances (AUC) for each of the sixteen classifiers on the entire testing set and every sub-population by permutations tests. Significant results at a confidence level of 0.05 were found for race and languages. The observed differences and the corresponding p-values yielded from the permutation tests are shown in Table 3 and Table 4. Among all the racial groups, we observed significant decreases in the performances of most of the classifiers for Asian and Hispanic patients (Table 3). Interestingly, significant performance drops were observed when applying the classifiers for the group of patients that speak Spanish (Table 4). We put the results of the social determinants associated with very few to no significant findings in the Supplementary Table 2-4 in the supplementary information. For a further illustration of the disparities, we showed the pairwise comparison results in Table 5 and Table 6. Among all the pairs of the racial groups, discrepancies were observed between Asian and White, as well as Asian and other races in most of the classifiers. Significant differences were also observed between Asian and Black sepsis patients in a few classifiers. The disparities between patients speaking various languages were majorly detected between the English-speaking patients and the Spanish-speaking patients. The pairwise comparison results with very few to no significant findings in the Supplementary Table 5-7 in the supplementary information.

# Discussion

Currently, the "gold standard" for sepsis diagnosis is still absent. Among those available criteria, we observed different sensitivities in identifying patients. Meanwhile, we observed disparities in the proportions of population identified by each criteria across various social determinant groups. This brings us the concern that a universal diagnostic system might not work equally on each sub-population. By systematically examining the discrepancies, we hope to provide evidence for a more versatile detection system that takes the disparities in social determinants into consideration.

The discrepancies among subpopulations of social determinants groups hinder the performance of a machine learning model trained on the entire population. In a previous study, racial disparities [35] and region disparities [36] in sepsis-related mortality were revealed by retrospective studies. Prediction of mortality using machine learning has been well-discussed during recent years. However, more effort was devoted to improving the overall performances on the entire given population. While what was being less discussed was the fairness of applying trained machine learning algorithms on various groups of patients. It is by nature that patients are of various social status and it is essential not to underestimate such discrepancies. In this current study, we tested the performance fluctuations when applying the same trained model on patients from each social determinant groups and revealed statistically significant shifts in the performance. Even though the overall performance of a given classifier is descent, it should be kept in mind that there are still sub-populations not benefitting from the model as others. On the one hand, we hope such evidence provides a perspective on the impacts of social determinants for not only the medical society that is working diligently towards a fairer diagnostic method but also the artificial intelligence researchers trying to improve the predictive algorithms one more step towards clinically ready. Additionally, in future studies, we would take the interaction between features into consideration for a more thorough perspective.

## Conclusions

Disparities in social determinants were observed in the groups of sepsis patients identified by various currently available diagnostic criteria. The performance of risk prediction tasks for sepsis patients can be compromised when applying a universally trained model for each sub-population. To achieve more accurate identification, a more versatile diagnostic system for sepsis is in need to overcome the social determinant disparities of patients.

## List of abbreviations

MIMIC-III: Medical Information Mart for Intensive Care - III

AUC: area under the receiver operating characteristic curve

AI: artificial intelligence

EHR: electronic health record

ICU: intensive care unit

ICD-9: International Classification of Diseases, 9th version

CMS: Centers for Medicare & Medicaid Services

CDC: Center of Disease Control and Prevention

kNN: k-nearest neighbors

linearSVC: support vector machine with linear kernel

SGD: stochastic gradient descent

EN: elastic net

NB: naive Bayes

SVC: support vector machine

SOFA: sequential organ failure assessment

SIRS: systemic inflammatory response syndrome

# Declarations

*Ethics approval and consent to participate*

Not applicable.

*Consent to publish*

Not applicable.

*Availability of data and materials*

The datasets supporting the conclusions of this article are available in the freely accessible database MIMIC-III through [PhysioNet](PhysioNet).

*Competing interests*

Authors have no competing interests except for receiving funding from NIH.

*Funding*

Dr. Luo reports funding from R01 LM013337.

Hanyin Wang reports funding from UL1 TR001422.

Dr. Naidech reports funding from R01 NS110779 and U01 NS110772.

*Authors' Contributions*

Dr. Yuan Luo, Dr. Naidech, and Hanyin Wang conceptualized the study. Hanyin Wang carried out all the data analysis, tables, and figures. Yikuan Li contributed to the optimization of the data visualizations.

*Acknowledgements*

All those who meaningfully contributed to the manuscripts are listed as an author.

Tables

Table 1. Statistics of 5,783 sepsis patients.

| Social Determinants | Category | n | % sepsis population | In-hospital mortality | % in-hospital mortality | Training | Testing |
|---|---|---|---|---|---|---|---|
| Race | Asian | 179 | 3.10% | 26 | 14.53% | 129 | 50 |
| | Black or African American | 501 | 8.66% | 52 | 10.38% | 348 | 153 |
| | Hispanic or Latino | 188 | 3.25% | 18 | 9.57% | 132 | 56 |
| | Other | 714 | 12.35% | 165 | 23.11% | 527 | 187 |
| | White | 4201 | 72.64% | 575 | 13.69% | 2912 | 1289 |
| Gender | Female | 2562 | 44.30% | 384 | 14.99% | 1798 | 764 |
| | Male | 3221 | 55.70% | 452 | 14.03% | 2250 | 971 |
| Marital status | Separated | 398 | 6.88% | 52 | 13.07% | 287 | 111 |
| | Significant other | 2559 | 44.25% | 363 | 14.19% | 1788 | 771 |
| | Single | 1638 | 28.32% | 174 | 10.62% | 1157 | 481 |
| | Unknown | 332 | 5.74% | 102 | 30.72% | 248 | 84 |
| | Widowed | 856 | 14.80% | 145 | 16.94% | 568 | 288 |
| Insurance type | Government | 166 | 2.87% | 13 | 7.83% | 115 | 51 |
| | Medicaid | 570 | 9.86% | 67 | 11.75% | 395 | 175 |
| | Medicare | 3358 | 58.07% | 560 | 16.68% | 2335 | 1023 |
| | Private | 1639 | 28.34% | 185 | 11.29% | 1168 | 471 |
| | Self-pay | 50 | 0.86% | 11 | 22.00% | 35 | 15 |
| Language | English | 5167 | 89.35% | 727 | 14.07% | 3631 | 1536 |
| | Other | 499 | 8.63% | 94 | 18.84% | 339 | 160 |
| | Spanish | 117 | 2.02% | 15 | 12.82% | 78 | 39 |

*Categories of each social determinant are ranked alphabetically; n: number of sepsis patients in the category; % sepsis population: percentage of the number of sepsis patients in the category among the 5,783 sepsis patients; In-hospital mortality: number of patients in the category deceased in-hospital; % in-hospital mortality: percentage of patients in the category deceased in-hospital; Training: number of patients of the given category that were assigned to the training set during train-test split; Testing: number of patients of the given category that were assigned to the test set during train-test split.*

Table 2. Detailed performances on the entire testing set.

|  | Accuracy | AUC | Precision | Recall | F1_binary | F1_macro | Specificity |
|---|---|---|---|---|---|---|---|
| **Ridge classifier** | 0.6790 | 0.7774 | 0.2682 | 0.7052 | 0.3886 | 0.5855 | 0.6745 |
| **Perceptron** | 0.6720 | 0.7786 | 0.2634 | 0.7052 | 0.3835 | 0.5801 | 0.6664 |
| **Passive-aggressive** | 0.6841 | 0.7582 | 0.2733 | 0.7131 | 0.3951 | 0.5907 | 0.6792 |
| **kNN** | 0.7135 | 0.7299 | 0.2780 | 0.6135 | 0.3826 | 0.5981 | 0.7305 |
| **Random forest** | 0.7516 | 0.6459 | 0.2826 | 0.4661 | 0.3519 | 0.5991 | 0.7999 |
| **LinearSVC_L1** | 0.6749 | 0.7781 | 0.2654 | 0.7052 | 0.3856 | 0.5823 | 0.6698 |
| **LinearSVC_L2** | 0.6784 | 0.7777 | 0.2678 | 0.7052 | 0.3882 | 0.5850 | 0.6739 |
| **SGDClassifier_L1** | 0.6790 | 0.7759 | 0.2682 | 0.7052 | 0.3886 | 0.5855 | 0.6745 |
| **SGDClassifier_L2** | 0.6790 | 0.7749 | 0.2668 | 0.6972 | 0.3859 | 0.5843 | 0.6759 |
| **SGDClassifier_EN** | 0.6801 | 0.7753 | 0.2683 | 0.7012 | 0.3881 | 0.5858 | 0.6765 |
| **MultinomialNB** | 0.6392 | 0.7040 | 0.2348 | 0.6614 | 0.3466 | 0.5487 | 0.6354 |
| **BernoulliNB** | 0.3107 | 0.5724 | 0.1665 | 0.9402 | 0.2830 | 0.3096 | 0.2042 |
| **Logistic regression** | 0.6824 | 0.7761 | 0.2720 | 0.7131 | 0.3938 | 0.5893 | 0.6772 |
| **SVC_rbf** | 0.6847 | 0.7744 | 0.2702 | 0.6932 | 0.3888 | 0.5882 | 0.6833 |
| **SVC_poly** | 0.6749 | 0.7751 | 0.2654 | 0.7052 | 0.3856 | 0.5823 | 0.6698 |
| **SVC_sigmoid** | 0.6277 | 0.6873 | 0.2349 | 0.6972 | 0.3514 | 0.5451 | 0.6159 |

*F1 binary: F1 score for the positive class; F1_macro: macro-averaged F1 score; Passive-aggressive: passive-aggressive classifier; kNN: k-Nearest Neighbors; LinearSVC_L1 or _L2: support vector machine with linear kernel coupled with L1 or L2 regularization; SGDClassifier_L1 or _L2 or _EN: stochastic gradient descent with L1 or L2 or Elastic Net regularization; MultinomialNB: Multinomial naïve Bayes; BernoulliNB: Bernoulli naïve Bayes; SVC_rbf or _poly or _sigmoid: support vector machine with rbf kernel or polynomial kernel or sigmoid kernel.*

Table 3. Observed differences between the testing results and each race with p-values from permutation tests.

| | Asian | | Black or African American | | Hispanic or Latino | | Other | | White | |
|---|---|---|---|---|---|---|---|---|---|---|
| | Observed difference | p_val | Observed difference | p_val | Observed difference | p_val | Observed difference | p_val | Observed difference | p_val |
| **Ridge classifier** | -0.2812 | *0.009* | -0.0241 | *0.366* | -0.2208 | *0.038* | 0.0011 | *0.528* | 0.0175 | *0.286* |
| **Perceptron** | -0.2748 | *0.007* | 0.0078 | *0.448* | -0.2453 | *0.025* | -0.0026 | *0.502* | 0.0158 | *0.312* |
| **Passive-aggressive** | -0.3188 | *0.003* | -0.0075 | *0.464* | -0.1749 | *0.083* | 0.0159 | *0.381* | 0.0111 | *0.370* |
| **kNN** | -0.1314 | *0.141* | -0.0628 | *0.219* | -0.1865 | *0.069* | -0.0144 | *0.372* | 0.0214 | *0.245* |
| **Random forest** | -0.0834 | *0.207* | -0.0939 | *0.046* | -0.1226 | *0.112* | 0.0536 | *0.120* | 0.0056 | *0.429* |
| **LinearSVC_L1** | -0.2819 | *0.012* | -0.0172 | *0.410* | -0.2247 | *0.039* | 0.0003 | *0.481* | 0.0173 | *0.285* |
| **LinearSVC_L2** | -0.2815 | *0.009* | -0.0221 | *0.385* | -0.2211 | *0.045* | 0.0005 | *0.490* | 0.0175 | *0.294* |
| **SGDClassifier_L1** | -0.2872 | *0.008* | -0.0041 | *0.482* | -0.2159 | *0.044* | 0.0036 | *0.478* | 0.0184 | *0.266* |
| **SGDClassifier_L2** | -0.2900 | *0.008* | -0.0087 | *0.455* | -0.2182 | *0.039* | 0.0044 | *0.469* | 0.0191 | *0.263* |
| **SGDClassifier_EN** | -0.2905 | *0.010* | -0.0046 | *0.461* | -0.2186 | *0.058* | 0.0050 | *0.497* | 0.0181 | *0.300* |
| **MultinomialNB** | -0.2797 | *0.010* | 0.0671 | *0.182* | -0.2373 | *0.033* | 0.0051 | *0.484* | 0.0051 | *0.416* |
| **BernoulliNB** | -0.1974 | *0.025* | -0.0034 | *0.483* | 0.0476 | *0.331* | 0.0173 | *0.368* | 0.0012 | *0.490* |
| **Logistic regression** | -0.2875 | *0.010* | -0.0257 | *0.377* | -0.2061 | *0.054* | 0.0043 | *0.495* | 0.0174 | *0.273* |
| **SVC_rbf** | -0.3085 | *0.005* | 0.0042 | *0.480* | -0.2311 | *0.031* | -0.0176 | *0.383* | 0.0175 | *0.275* |
| **SVC_poly** | -0.2978 | *0.006* | 0.0027 | *0.483* | -0.2751 | *0.017* | 0.0087 | *0.431* | 0.0154 | *0.287* |
| **SVC_sigmoid** | -0.1343 | *0.144* | -0.0941 | *0.083* | -0.0606 | *0.332* | -0.0099 | *0.455* | 0.0208 | *0.247* |

*Observe difference: observed difference in AUC when compared with the performance on the entire testing set; p_val: p-value, p-values less than or equal to 0.05 were highlighted; Passive-aggressive: passive-aggressive classifier; kNN: k-Nearest Neighbors; LinearSVC_L1 or _L2: support vector machine with linear kernel coupled with L1 or L2 regularization; SGDClassifier_L1 or _L2 or _EN: stochastic gradient descent with L1 or L2 or Elastic Net regularization; MultinomialNB: Multinomial naïve Bayes; BernoulliNB: Bernoulli naïve Bayes; SVC_rbf or _poly or _sigmoid: support vector machine with rbf kernel or polynomial kernel or sigmoid kernel.*

Table 4. Observed differences between the testing results and each language with p-values from permutation tests.

|  | English | | Other | | Spanish | |
| --- | --- | --- | --- | --- | --- | --- |
|  | Observed difference | p_val | Observed difference | p_val | Observed difference | p_val |
| **Ridge classifier** | 0.0154 | 0.279 | -0.0760 | 0.107 | -0.3422 | *0.012* |
| **Perceptron** | 0.0182 | 0.252 | -0.0916 | 0.053 | -0.3551 | *0.004* |
| **Passive-aggressive** | 0.0122 | 0.301 | -0.0555 | 0.172 | -0.2288 | 0.053 |
| **kNN** | 0.0166 | 0.263 | -0.0768 | 0.102 | -0.3063 | *0.017* |
| **Random forest** | 0.0037 | 0.409 | -0.0057 | 0.489 | -0.2342 | *0.002* |
| **LinearSVC_L1** | 0.0160 | 0.299 | -0.0772 | 0.102 | -0.3428 | *0.003* |
| **LinearSVC_L2** | 0.0156 | 0.297 | -0.0763 | 0.121 | -0.3424 | *0.007* |
| **SGDClassifier_L1** | 0.0184 | 0.246 | -0.0783 | 0.093 | -0.3347 | *0.008* |
| **SGDClassifier_L2** | 0.0187 | 0.269 | -0.0752 | 0.107 | -0.3396 | *0.004* |
| **SGDClassifier_EN** | 0.0181 | 0.259 | -0.0760 | 0.105 | -0.3283 | *0.006* |
| **MultinomialNB** | 0.0221 | 0.224 | -0.1210 | *0.021* | -0.2746 | *0.031* |
| **BernoulliNB** | 0.0076 | 0.389 | -0.0621 | 0.082 | 0.0306 | 0.422 |
| **Logistic regression** | 0.0145 | 0.293 | -0.0703 | 0.125 | -0.3173 | *0.014* |
| **SVC_rbf** | 0.0159 | 0.306 | -0.0825 | 0.080 | -0.3332 | *0.012* |
| **SVC_poly** | 0.0176 | 0.275 | -0.0860 | 0.079 | -0.3633 | *0.002* |
| **SVC_sigmoid** | -0.0030 | 0.454 | 0.0341 | 0.288 | -0.1814 | 0.089 |

*Observe difference: observed difference in AUC when compared with the performance on the entire testing set; p_val: p-value, p-values less than or equal to 0.05 were highlighted; Passive-aggressive: passive-aggressive classifier; kNN: k-Nearest Neighbors; LinearSVC_L1 or _L2: support vector machine with linear kernel coupled with L1 or L2 regularization; SGDClassifier_L1 or _L2 or _EN: stochastic gradient descent with L1 or L2 or Elastic Net regularization; MultinomialNB: Multinomial naïve Bayes; BernoulliNB: Bernoulli naïve Bayes; SVC_rbf or _poly or _sigmoid: support vector machine with rbf kernel or polynomial kernel or sigmoid kernel.*

Table 5. Pairwise comparisons among different racial groups.

| | Asian v.s. Black or African American | | Asian v.s. Hispanic or Latino | | Asian v.s. Other | | Asian v.s. White | | Black or African American v.s. Hispanic or Latino | |
|---|---|---|---|---|---|---|---|---|---|---|
| | Observed difference | p_val | Observed difference | p_val | Observed difference | p_val | Observed difference | p_val | Observed difference | p_val |
| **Ridge classifier** | 0.2572 | 0.074 | 0.0605 | 0.738 | 0.2824 | *0.033* | 0.2988 | *0.018* | -0.1967 | 0.189 |
| **Perceptron** | 0.2827 | *0.042* | 0.0295 | 0.883 | 0.2722 | 0.051 | 0.2906 | *0.021* | -0.2531 | 0.081 |
| **Passive-aggressive** | 0.3114 | *0.045* | 0.1439 | 0.432 | 0.3348 | *0.018* | 0.3299 | *0.008* | -0.1674 | 0.238 |
| **kNN** | 0.0686 | 0.647 | -0.0552 | 0.763 | 0.1170 | 0.380 | 0.1528 | 0.224 | -0.1238 | 0.413 |
| **Random forest** | -0.0104 | 0.916 | -0.0392 | 0.715 | 0.1370 | 0.211 | 0.0890 | 0.372 | -0.0287 | 0.781 |
| **LinearSVC_L1** | 0.2647 | 0.075 | 0.0571 | 0.756 | 0.2822 | *0.043* | 0.2991 | *0.020* | -0.2076 | 0.156 |
| **LinearSVC_L2** | 0.2594 | 0.084 | 0.0605 | 0.752 | 0.2820 | *0.042* | 0.2990 | *0.019* | -0.1990 | 0.179 |
| **SGDClassifier_L1** | 0.2832 | 0.052 | 0.0714 | 0.668 | 0.2908 | *0.036* | 0.3057 | *0.022* | -0.2118 | 0.136 |
| **SGDClassifier_L2** | 0.2813 | *0.050* | 0.0718 | 0.692 | 0.2944 | *0.019* | 0.3091 | *0.015* | -0.2095 | 0.151 |
| **SGDClassifier_EN** | 0.2858 | 0.058 | 0.0718 | 0.706 | 0.2954 | *0.035* | 0.3086 | *0.015* | -0.2140 | 0.142 |
| **MultinomialNB** | 0.3468 | *0.013* | 0.0424 | 0.800 | 0.2848 | *0.035* | 0.2848 | *0.029* | -0.3044 | *0.032* |
| **BernoulliNB** | 0.1940 | *0.043* | 0.2450 | 0.068 | 0.2147 | *0.015* | 0.1986 | *0.021* | 0.0510 | 0.609 |
| **Logistic regression** | 0.2617 | 0.082 | 0.0814 | 0.620 | 0.2918 | *0.037* | 0.3049 | *0.019* | -0.1804 | 0.198 |
| **SVC_rbf** | 0.3127 | *0.025* | 0.0774 | 0.653 | 0.2909 | *0.030* | 0.3259 | *0.013* | -0.2352 | 0.093 |
| **SVC_poly** | 0.3005 | *0.024* | 0.0227 | 0.889 | 0.3066 | *0.025* | 0.3132 | *0.016* | -0.2778 | 0.056 |
| **SVC_sigmoid** | 0.0402 | 0.780 | 0.0736 | 0.666 | 0.1244 | 0.375 | 0.1551 | 0.235 | 0.0334 | 0.796 |

| | Black or African American v.s. Other | | Black or African American v.s. White | | Hispanic or Latino v.s. Other | | Hispanic or Latino v.s. White | | Other v.s. White | |
|---|---|---|---|---|---|---|---|---|---|---|
| | Observed difference | p_val | Observed difference | p_val | Observed difference | p_val | Observed difference | p_val | Observed difference | p_val |
| **Ridge classifier** | 0.0252 | 0.781 | 0.0416 | 0.557 | 0.2219 | 0.103 | 0.2383 | 0.055 | 0.0164 | 0.783 |
| **Perceptron** | -0.0104 | 0.930 | 0.0080 | 0.926 | 0.2427 | 0.076 | 0.2611 | *0.032* | 0.0184 | 0.747 |
| **Passive-aggressive** | 0.0234 | 0.764 | 0.0186 | 0.791 | 0.1908 | 0.167 | 0.1860 | 0.134 | -0.0048 | 0.931 |

| Classifier | Observe difference | p_val | Observe difference | p_val | Observe difference | p_val | Observe difference | p_val | Observe difference | p_val |
|---|---|---|---|---|---|---|---|---|---|---|
| kNN | 0.0484 | *0.564* | 0.0841 | *0.225* | 0.1721 | *0.189* | 0.2079 | *0.081* | 0.0358 | *0.537* |
| Random forest | 0.1474 | *0.029* | 0.0994 | *0.089* | 0.1762 | *0.101* | 0.1282 | *0.182* | -0.0480 | *0.278* |
| LinearSVC_L1 | 0.0175 | *0.832* | 0.0344 | *0.629* | 0.2251 | *0.106* | 0.2420 | *0.065* | 0.0170 | *0.764* |
| LinearSVC_L2 | 0.0226 | *0.792* | 0.0396 | *0.585* | 0.2216 | *0.088* | 0.2386 | *0.065* | 0.0170 | *0.756* |
| SGDClassifier_L1 | 0.0076 | *0.931* | 0.0225 | *0.753* | 0.2194 | *0.108* | 0.2343 | *0.075* | 0.0149 | *0.794* |
| SGDClassifier_L2 | 0.0131 | *0.882* | 0.0278 | *0.699* | 0.2226 | *0.080* | 0.2373 | *0.059* | 0.0147 | *0.786* |
| SGDClassifier_EN | 0.0096 | *0.932* | 0.0228 | *0.765* | 0.2236 | *0.088* | 0.2368 | *0.070* | 0.0132 | *0.830* |
| MultinomialNB | -0.0620 | *0.491* | -0.0620 | *0.425* | 0.2423 | *0.073* | 0.2424 | *0.053* | 0.0001 | *1.000* |
| BernoulliNB | 0.0207 | *0.702* | 0.0046 | *0.935* | -0.0303 | *0.764* | -0.0464 | *0.607* | -0.0161 | *0.650* |
| Logistic regression | 0.0301 | *0.705* | 0.0432 | *0.579* | 0.2104 | *0.130* | 0.2235 | *0.083* | 0.0131 | *0.827* |
| SVC_rbf | -0.0218 | *0.799* | 0.0133 | *0.860* | 0.2135 | *0.110* | 0.2485 | *0.047* | 0.0350 | *0.527* |
| SVC_poly | 0.0060 | *0.930* | 0.0127 | *0.848* | 0.2838 | *0.027* | 0.2905 | *0.019* | 0.0066 | *0.904* |
| SVC_sigmoid | 0.0841 | *0.286* | 0.1149 | *0.110* | 0.0507 | *0.727* | 0.0814 | *0.544* | 0.0307 | *0.584* |

*Observe difference: observed difference in AUC when comparing the performance between the sub-populations; p_val: p-value, p-values less than or equal to 0.05 were highlighted; Passive-aggressive: passive-aggressive classifier; kNN: k-Nearest Neighbors; LinearSVC_L1 or _L2: support vector machine with linear kernel coupled with L1 or L2 regularization; SGDClassifier_L1 or _L2 or _EN: stochastic gradient descent with L1 or L2 or Elastic Net regularization; MultinomialNB: Multinomial naïve Bayes; BernoulliNB: Bernoulli naïve Bayes; SVC_rbf or _poly or _sigmoid: support vector machine with rbf kernel or polynomial kernel or sigmoid kernel.*

Table 6. Pairwise comparisons among different language groups.

| | English v.s. Other | | English v.s. Spanish | | Other v.s. Spanish | |
|---|---|---|---|---|---|---|
| | Observed difference | p_val | Observed difference | p_val | Observed difference | p_val |
| **Ridge classifier** | -0.0915 | 0.135 | -0.3576 | *0.008* | -0.2661 | 0.081 |
| **Perceptron** | -0.1098 | 0.070 | -0.3733 | *0.005* | -0.2635 | 0.095 |
| **Passive-aggressive** | -0.0677 | 0.266 | -0.2410 | 0.087 | -0.1733 | 0.281 |
| **kNN** | -0.0934 | 0.159 | -0.3230 | *0.021* | -0.2295 | 0.121 |
| **Random forest** | -0.0094 | 0.833 | -0.2379 | *0.023* | -0.2285 | 0.064 |
| **LinearSVC_L1** | -0.0931 | 0.132 | -0.3587 | *0.007* | -0.2656 | 0.097 |
| **LinearSVC_L2** | -0.0919 | 0.135 | -0.3580 | *0.008* | -0.2661 | 0.095 |
| **SGDClassifier_L1** | -0.0967 | 0.113 | -0.3531 | *0.015* | -0.2564 | 0.097 |
| **SGDClassifier_L2** | -0.0939 | 0.143 | -0.3583 | *0.009* | -0.2643 | 0.078 |
| **SGDClassifier_EN** | -0.0940 | 0.136 | -0.3463 | *0.009* | -0.2523 | 0.093 |
| **MultinomialNB** | -0.1432 | *0.017* | -0.2967 | *0.034* | -0.1535 | 0.295 |
| **BernoulliNB** | -0.0697 | 0.091 | 0.0230 | 0.818 | 0.0927 | 0.397 |
| **Logistic regression** | -0.0849 | 0.174 | -0.3318 | *0.025* | -0.2469 | 0.093 |
| **SVC_rbf** | -0.0984 | 0.112 | -0.3491 | *0.013* | -0.2507 | 0.104 |
| **SVC_poly** | -0.1035 | 0.100 | -0.3809 | *0.009* | -0.2773 | 0.072 |
| **SVC_sigmoid** | 0.0372 | 0.544 | -0.1784 | 0.203 | -0.2155 | 0.151 |

*Observe difference: observed difference in AUC when comparing the performance between the sub-populations; p_val: p-value, p-values less than or equal to 0.05 were highlighted; Passive-aggressive: passive-aggressive classifier; kNN: k-Nearest Neighbors; LinearSVC_L1 or _L2: support vector machine with linear kernel coupled with L1 or L2 regularization; SGDClassifier_L1 or _L2 or _EN: stochastic gradient descent with L1 or L2 or Elastic Net regularization; MultinomialNB: Multinomial naïve Bayes; BernoulliNB: Bernoulli naïve Bayes; SVC_rbf or _poly or _sigmoid: support vector machine with rbf kernel or polynomial kernel or sigmoid kernel.*